\documentclass[final,5p,times,twocolumn]{elsarticle}

\usepackage{amssymb}
\usepackage{amsmath}
\usepackage{amsfonts}
\usepackage{amsthm}
\usepackage{mathrsfs}
\usepackage[table]{xcolor} % Loaded with table option for \rowcolors
\usepackage{textcomp}
\usepackage{booktabs}
\usepackage{algorithm}
\usepackage{algorithmicx}
\usepackage{algpseudocode}
\usepackage{listings}
\usepackage{graphicx}
\usepackage{multirow}

\usepackage{adjustbox}
\usepackage{tabularx}
\usepackage{tikz}
\usepackage{url}
\usepackage{hyperref}
\usepackage{colortbl}
\usepackage{placeins}
\usepackage{makecell}
\usepackage{pgfplots}
\usepackage{subcaption}
\usepackage{titlesec}

\definecolor{myblue}{RGB}{0, 70, 200}
\definecolor{mygreen}{RGB}{0, 150, 30}
\definecolor{mygray}{RGB}{180, 180, 180}
\definecolor{localorange}{HTML}{CC7A00}
\definecolor{globalgreen}{HTML}{1A7F37}
\definecolor{frozenblue}{HTML}{2563EB}

\definecolor{cclipmcm}{RGB}{150,150,150}
\definecolor{cclipgl}{RGB}{120,120,120}
\definecolor{ccoopmcm}{RGB}{227,119,194}
\definecolor{ccoopgl}{RGB}{199,74,162}
\definecolor{cplot}{RGB}{31,119,180}
\definecolor{clocoopm}{RGB}{44,160,44}
\definecolor{clocoopgl}{RGB}{30,110,30}
\definecolor{cpromptsrc}{RGB}{255,127,14}
\definecolor{clsncoop}{RGB}{148,103,189}
\definecolor{clsncocoop}{RGB}{107,70,154}
\definecolor{cgallop}{RGB}{214,39,40}
\definecolor{cours}{RGB}{0,114,178}
\definecolor{coursnol}{RGB}{230,159,0}

\tikzset{
  baseDot/.style = {fill=mygray},
  ourBlue/.style = {fill=myblue, text=myblue},
  ourGreen/.style = {fill=mygreen, draw=mygreen, text=mygreen},
  pics/snowbadge/.style={
    code={
      \begin{scope}[scale=1]
        \fill[white] (0,0) circle (1.1mm);
        \draw[frozenblue, line width=0.5pt] (0,0) circle (1.1mm);
        \foreach \a in {0,60,...,300}{
          \draw[frozenblue, line width=0.6pt] (0,0) -- (\a:1.0mm);
          \draw[frozenblue, line width=0.5pt] (\a:0.65mm) -- (\a+30:0.38mm);
          \draw[frozenblue, line width=0.5pt] (\a:0.65mm) -- (\a-30:0.38mm);
        }
      \end{scope}
    }
  },
  pics/firebadge/.style={
    code={
      \begin{scope}[x=3mm,y=1.5mm]
        \path[draw=orange!60!red, line width=0.16pt, fill=orange!80!red]
          (0,0) .. controls (0.45,0.9) and (0.35,1.2) .. (0,1.6)
                .. controls (-0.35,1.2) and (-0.45,0.9) .. (0,0) -- cycle;
        \path[fill=yellow!85!white, draw=none]
          (0,0.55) .. controls (0.18,0.95) .. (0,1.25)
                    .. controls (-0.18,0.95) .. (0,0.55) -- cycle;
      \end{scope}
    }
  },
  arr/.style={-{Stealth[length=2.6mm]}, line width=1pt},
  common/.style={line width=0.3pt},
  mclipmcm/.style={common, cclipmcm, mark=o, mark size=1.2pt},
  mclipgl/.style={common, cclipgl, mark=*, mark size=1.0pt},
  mcoopmcm/.style={common, ccoopmcm, mark=triangle*, mark size=1.2pt},
  mcoopgl/.style={common, ccoopgl, mark=triangle, mark size=1.3pt},
  mplot/.style={common, cplot, mark=square*, mark size=1.1pt},
  mlocoopm/.style={common, clocoopm, mark=diamond*, mark size=1.2pt},
  mlocoopgl/.style={common, clocoopgl, mark=diamond, mark size=1.3pt},
  mprompt/.style={common, cpromptsrc, mark=asterisk, mark size=1.5pt},
  mlsncoop/.style={common, clsncoop, mark=oplus*, mark size=1.3pt},
  mlsncocoop/.style={common, clsncocoop, mark=oplus, mark size=1.4pt},
  mgallop/.style={common, cgallop, mark=otimes*, mark size=1.1pt},
  mours/.style={cours, mark=star, mark size=2.6pt, line width=0.5pt, fill=white},
  moursnol/.style={coursnol, mark=star, mark size=2.6pt, line width=0.5pt},
}
\usetikzlibrary{arrows.meta,positioning,calc,decorations.pathreplacing,fit,3d}

\begin{document}

\begin{frontmatter}

%% Title
%\title{SOT-GLP: Sparse Optimal Transport Guided Local-Global Prompt Learning}
\title{Local-Global Prompt Learning via Sparse Optimal Transport}
%% Authors
\author[1]{Deniz Kizaroğlu\corref{cor1}}
\ead{deniz.kizaroglu@metu.edu.tr}

\author[2]{Ülkü Tuncer Küçüktaş}
\ead{24830604018@gazi.edu.tr}

\author[1]{Emre Çakmakyurdu}
\ead{emre.cakmakyurdu@metu.edu.tr}

\author[1]{Alptekin Temizel}
\ead{atemizel@metu.edu.tr}

%% Affiliations
\affiliation[1]{organization={Graduate School of Informatics, Middle East Technical University},
            city={Ankara},
            country={Turkey}}

\affiliation[2]{organization={Department of Electrical and Electronics Engineering, Gazi University},
            city={Ankara},
            country={Turkey}}

\cortext[cor1]{Corresponding author}

%% Abstract
\begin{abstract}
Few-shot adaptation of vision-language models (VLMs) like CLIP typically relies on learning textual prompts matched to global image embeddings. Recent works extend this paradigm by incorporating local image--text alignment to capture fine-grained visual cues, yet these approaches often select local regions independently for each prompt, leading to redundant local feature usage and prompt overlap. We propose SOT-GLP, which introduces a shared sparse patch support and balanced optimal transport allocation to explicitly partition salient visual regions among class-specific local prompts, while preserving global alignment. Our method learns shared global prompts and class-specific local prompts. The global branch maintains standard image-text matching for robust category-level alignment. The local branch constructs a class-conditioned sparse patch set using V–V attention and aligns it to multiple class-specific prompts via balanced entropic optimal transport, yielding a soft partition of patches that prevents prompt overlap and collapse. We evaluate our method on two complementary objectives: (i) few-shot classification accuracy on 11 standard benchmarks, and (ii) out-of-distribution (OOD) detection. On the standard 11-dataset benchmark with 16-shot ViT-B/16, SOT-GLP achieves 85.1\% average accuracy, outperforming prior prompt-learning methods. We identify a distinct accuracy-robustness trade-off in prompt learning: while learnable projections optimize in-distribution fit, they alter the foundational feature space. We demonstrate that a projection-free local alignment preserves the native geometry of the CLIP manifold, yielding state-of-the-art OOD detection performance (94.2\% AUC) that surpasses fully adapted models. Implementation available at: \url{https://github.com/Deniz2304988/SOT-GLP}
\end{abstract}

%% Keywords

\end{frontmatter}

%% Main Text

\section{Introduction}
Large-scale vision–language models (VLMs) such as CLIP, ALIGN, and BLIP \cite{radford2021learning,jia2021scaling,li2022blip} have shown that pairing images with natural-language supervision is a powerful alternative to fixed-category supervision, achieving strong zero- and few-shot performance across recognition and retrieval benchmarks. By learning a shared embedding space, novel categories (zero-shot) can be recognized by encoding category names as text prompts, then classifying images by selecting the text embedding with highest similarity to the image embedding. In practice, however, this ability remains limited by two main factors: the quality of the text prompts and the image encoder's ability to capture fine-grained visual structure.

Prompt learning methods, such as CoOp \cite{zhou2022learning}, address the prompt quality issue by replacing hand-written templates with learnable textual prompts adapted on small labeled sets. While this substantially reduces the domain gap, most existing approaches rely on matching a single global image embedding (typically the [CLS] token) to a global text embedding. This whole-image objective averages over all spatial regions, discarding fine-grained local features—such as object parts, texture patterns, and spatial configurations—that are critical for discriminating similar categories and detecting out-of-distribution (OOD) samples. Furthermore, approaches that try to use local features perform independent per-prompt patch selection, which allows multiple prompts to attend to overlapping regions and limits effective specialization. We address the problem of multi-prompt local evidence allocation, where multiple class-specific prompts must specialize on distinct visual evidence extracted from a shared set of salient patches.

\begin{figure*}[t]
\centering
\begin{adjustbox}{max totalsize={\textwidth}{1.2\textheight},center}
\begin{tikzpicture}[x=6mm,y=6mm,line join=round,line cap=round]

% ---- styles ----
\tikzset{
  cell/.style ={draw,rounded corners=1.2pt,line width=0.6pt,fill=white},
  cellB/.style={cell,fill=green!22},
  cellR/.style={cell,fill=green!22}
}

\begin{scope}[shift={(-32.2,0.0)}] 
  \tikzset{
  block/.style={draw=blue!60!black, rounded corners=2pt,
                fill=blue!8!white, minimum width=19mm, minimum height=8mm,
                line width=0.9pt, font=\sffamily\small, align=center},
blockS/.style={block, minimum height=6.8mm},
  arr/.style={-{Stealth[length=2.6mm]}, line width=1pt},
  dotsstyle/.style={font=\Large\bfseries, text=black, fill=white,
                    inner sep=1.5pt, rounded corners=2pt}
    }

  % --- left trio ---
  \node[block]  (vit1) at (0,1.8) {\textbf{VIT Layer}};
  \node[block]  (vit2) [right=12mm of vit1] {\textbf{VIT Layer}};
  \node[blockS] (att1) at (0,0)   {\textbf{V-V attn}};
  \node[blockS] (att2) [right=12mm of att1] {\textbf{V-V attn}};

  % connections (left trio)
  \draw[arr] (vit1.east)--(vit2.west);
  \draw[arr] (att1.east)--(att2.west);
  \draw[arr] (att1.east) -- ++(3mm,0) |- (vit2.west);

  % ellipses between groups
  \node[dotsstyle, right=2mm of vit2] (dotsT) {$\cdots$};
  \node[dotsstyle, right=2mm of att2] (dotsB) {$\cdots$};

  % --- right trio ---
  \node[block]  (vit4) [right=2mm of dotsT] {\textbf{VIT Layer}};
  \node[block]  (vit5) [right=12mm of vit4] {\textbf{VIT Layer}};
  \node[blockS] (att4) [right=2mm of dotsB] {\textbf{V-V attn}};
  \node[blockS] (att5) [right=12mm of att4] {\textbf{V-V attn}};

  % connections (right trio)
  \draw[arr] (vit4.east)--(vit5.west);
  \draw[arr] (att4.east)--(att5.west);
  \draw[arr] (att4.east) -- ++(3mm,0) |- (vit5.west);

  % outputs
  \node[anchor=west,font=\sffamily\bfseries] (ZORI) at ([xshift=5mm]vit5.east) {$\mathbf{Z}_{\mathrm{global}}$};
  \node[anchor=west,font=\sffamily\bfseries] (Z_local)   at ([xshift=5mm]att5.east) {$\mathbf{Z}_{\mathrm{local}}$};
\end{scope}

\draw[-{Stealth[length=2.6mm]}, line width=1pt, draw=globalgreen]
  (vit5.east) -- (ZORI.west);

\draw[-{Stealth[length=2.6mm]}, line width=1pt, draw=localorange]
  (att5.east) -- (Z_local.west);

\node[draw=localorange, rounded corners=2pt, fill=white, line width=0.9pt,
      font=\sffamily\scriptsize\bfseries, align=center, inner sep=1.5pt,minimum height=8mm, minimum width=14mm,
      right=6mm of Z_local] (projBlock) {Local\\Proj.};
      
% geometry knobs
\def\xs{0.95}   % spacing x
\def\ys{0.95}   % spacing y
\def\rot{90}    % rotation angle (approx your ref)
\def\shear{0.135}% x-slant; controls the "lean"

% helper: rectangle centered at (i*xs, j*ys)
\newcommand{\rcell}[4]{% i j halfsize style
  \draw[#4] (#1*\xs-#3, #2*\ys-#3) rectangle ++(2*#3, 2*#3);
}

% ----- BLUE grid (back, upper-right) -----
\begin{scope}[shift={(2.5,0.55)}, rotate=\rot, xslant=\shear,  local bounding box=simgrid]
  \def\h{0.40}
  \foreach \i in {0,...,4}{
    \foreach \j in {0,...,3}{
      % highlight 2x2 upper-right
      \ifnum\i>1\relax
        \ifnum\j>0\relax
          \rcell{\i}{\j}{\h}{cellB}
        \else \rcell{\i}{\j}{\h}{cell} \fi
      \else \rcell{\i}{\j}{\h}{cell} \fi
    }
  }
\end{scope}

% ----- RED grid (front, lower-left) -----
\begin{scope}[shift={(0.15,-0.25)}, rotate=\rot, xslant=\shear]
  \def\h{0.40}
  \foreach \i in {0,...,4}{
    \foreach \j in {0,...,3}{
      % highlight 2x2 lower-left
      \ifnum\i<2\relax
        \ifnum\j<2\relax
          \rcell{\i}{\j}{\h}{cellR}
        \else \rcell{\i}{\j}{\h}{cell} \fi
      \else \rcell{\i}{\j}{\h}{cell} \fi
    }
  }
\end{scope}

\node(simtopk)[align=center, anchor=south, font=\Large]
  at (0.5,5.5) {$\mathrm{sim}_{\text{top-}k}\!\big(Z_{\text{local}}, L_c\big)$};

\draw[-{Stealth[length=2.6mm]}, line width=1pt, draw=localorange]
  (Z_local.east) -- (projBlock.west);

\draw[-{Stealth[length=3mm]}, line width=1.2pt, draw=localorange]
  let \p1 = (projBlock.east) in
  (\p1) -- (\x1+12mm,\y1) -- (\x1+12mm,2.0) -- (-3.5,2.0);

\tikzset{
  encbox/.style={draw=frozenblue, rounded corners=2pt, fill=black!5,
                 line width=0.9pt, minimum width=32mm, minimum height=13mm,
                 font=\sffamily\bfseries\large, align=center},
  ptokenG/.style={draw=green!60!black, rounded corners=1pt, line width=0.7pt,
                 minimum width=4.8mm, minimum height=4.8mm, fill=white},
  ptokenL/.style={draw=orange!85!black, rounded corners=1pt, line width=0.7pt,
                 minimum width=4.8mm, minimum height=4.8mm, fill=orange!20}
}

\begin{scope}[shift={(-30.0,10.3)}] % move the whole top panel; tweak if needed

  % ----- Global prompts: NO class-specific learnables -----
  \foreach \i in {0,...,3} {
    \draw[ptokenG] (-2.2, -\i*0.9) rectangle ++(0.75,0.75);
    \node[anchor=west] at (-1.25, -\i*0.9+0.375) {$+\langle \text{class} \rangle$};
  }
  \node[align=center] at (-1.0, -4*0.9 - 0.55) {$\mathcal{P}_g$\\\footnotesize global prompts \\\footnotesize (non-class specific) };

  \tikzset{
  classchip/.style={draw=black!35, rounded corners=1pt, fill=black!5,
                    minimum width=7mm, minimum height=4.8mm,
                    font=\sffamily\scriptsize, inner sep=1pt},
    }
  \foreach \i/\cl in {0/{$c_1$},1/{$c_2$},2/{$c_3$},3/{$c_4$}} {
  % class label chip
  \node[classchip, anchor=east] at (2.95, -\i*0.9+0.375) {\cl};
  % "+ <class>" text
  \node[anchor=west] at (3.10, -\i*0.9+0.375) {$+\langle \text{class} \rangle$};
}
  
  \node[align=center] at (3.70, -4*0.9 - 0.55) {$\mathcal{P}_l$\\\footnotesize local prompts \\\footnotesize (class specific) };

  % ----- Frozen Text Encoder -----
  \node[encbox, anchor=west] (textencTop) at (10.6, -0.5) {Text\\Encoder};

  % ----- Arrows into encoder (global=green, local=orange) -----
  \draw[-{Stealth[length=2.6mm]}, line width=1.0pt, draw=globalgreen]
  (-0.6,0.9) -- (-0.6,3.0) -- (9.0,3.0) |- (textencTop.north west);

\draw[-{Stealth[length=2.6mm]}, line width=1.0pt, draw=localorange]
  (3.4,0.9)  -- (3.4,1.5)  -- (7.5,1.5) |- (textencTop.south west);

  \tikzset{
  tokG/.style={draw=globalgreen!80!black, fill=globalgreen!12,
               rounded corners=1.2pt, line width=0.9pt,
               minimum width=5.2mm, minimum height=5.2mm},
  tokL/.style={draw=localorange!85!black, fill=localorange!20,
               rounded corners=1.2pt, line width=0.9pt,
               minimum width=5.2mm, minimum height=5.2mm},
}

% --- rows of tokens to the right of the encoder ---
% anchor points for rows (use 'positioning'; no calc needed)
\node[coordinate, right=7mm of textencTop, yshift=4.8mm] (g0) {};
\node[coordinate, right=7mm of textencTop, yshift=-4.8mm] (l0) {};

% global (green) row
\node[tokG, right=0mm of g0] (g1) {};
\node[tokG, right=0.8mm of g1] (g2) {};
\node[tokG, right=0.8mm of g2] (g3) {};
\node[tokG, right=0.8mm of g3] (g4) {};

% local (orange) row
\node[tokL, right=0mm of l0] (l1) {};
\node[tokL, right=0.8mm of l1] (l2) {};
\node[tokL, right=0.8mm of l2] (l3) {};
\node[tokL, right=0.8mm of l3] (l4) {};

% short arrows from encoder to rows
\draw[-{Stealth[length=2.2mm]}, line width=1.0pt, draw=globalgreen]
  (textencTop.east) ++(0,4.8mm) -- (g1.west);
\draw[-{Stealth[length=2.2mm]}, line width=1.0pt, draw=localorange]
  (textencTop.east) ++(0,-4.8mm) -- (l1.west);

\node[draw, rounded corners=2pt, fill=black!3, line width=0.9pt,
        minimum width=28mm, minimum height=10mm,
        font=\sffamily\bfseries] (globloss) [right=50mm of g4] {Global Loss};

\draw[-{Stealth[length=2.2mm]}, line width=1.0pt, draw=globalgreen]
  (g4.east) -- ++(2mm,0) |- (globloss.north west);

% Dashed arrow from Z_global ([CLS]) to Global Loss
\draw[dashed, -{Stealth[length=2.2mm]}, line width=1.0pt, draw=globalgreen]
  let \p1 = (ZORI.north), \p2 = (globloss.south west) in
  (\p1) -- (\x1,\y1+20mm)                 % up
        -- (\x2-30mm,\y1+20mm)                  % right
        -- (\x2-30mm,\y2)                   % up
        node[above, font=\scriptsize\sffamily, text=globalgreen!70!black]{[CLS]}
        -- (\p2);                        % right into box
  
\draw[-{Stealth[length=2.2mm]}, line width=1.0pt, draw=localorange]
  let \p1 = (l4.east), \p2 = (simtopk.north) in
  (\p1) -- node[above=1pt, pos=0.1, font=\sffamily\small, text=localorange]{\textbf{Lc}} (\x2,\y1) -- (\p2);

\coordinate (simgridRight) at (simgrid.east);
\coordinate (otOrigin)       at ([xshift=26mm]simgridRight);

% --- 3D Optimal Local Loss panel (scaled to fit) ---
\begin{scope}[
  shift={(otOrigin)}, scale=0.1,
  x={(2.4cm,0.55cm)},    % less wide
  y={(0.70cm,-0.75cm)}, % taller projection
  z={(0cm,3.6cm)},        % z separation
  line cap=round, >=latex
]
  \def\L{18}
  \def\N{7}

  % Image plane with grid
  \begin{scope}[canvas is xy plane at z=0]
    \node[anchor=center, inner sep=0pt, transform shape, yscale=-1]
      at ({\L/2},{\L/2},0)
      {\includegraphics[width=\L cm,height=\L cm]{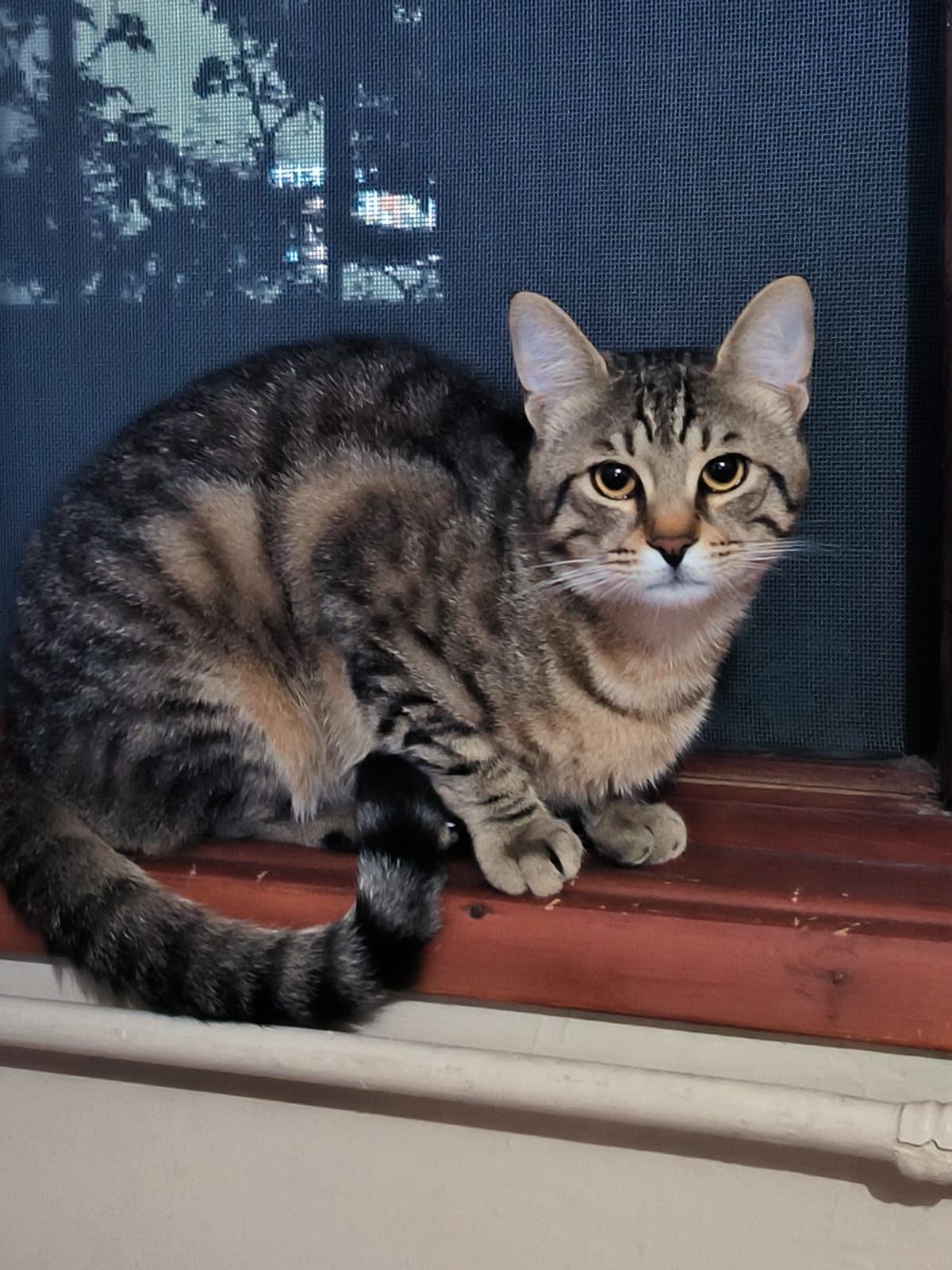}};
    \foreach \i in {0,...,\N}{
      \draw[blue!60,line width=0.3mm] (0,{\i*\L/\N},0) -- (\L,{\i*\L/\N},0);
      \draw[blue!60,line width=0.3mm] ({\i*\L/\N},0,0) -- ({\i*\L/\N},\L,0);
    }
  \end{scope}

  % Token centers
  \coordinate (T1) at (-1,10,5);
  \coordinate (T2) at ( 2,13,5);
  \coordinate (T3) at (14, 8,5);
  \coordinate (T4) at (12,-2,4);

  % Token squares
  \def\tokensize{3}
  \begin{scope}[canvas is xy plane at z=5]
    \fill[pink!70]             ({-1-\tokensize/2},{10-\tokensize/2})
                              rectangle ++({\tokensize},{\tokensize});
    \fill[blue!60]             ({ 2-\tokensize/2},{13-\tokensize/2})
                              rectangle ++({\tokensize},{\tokensize});
    \fill[green!70!black]      ({14-\tokensize/2},{ 8-\tokensize/2})
                              rectangle ++({\tokensize},{\tokensize});
  \end{scope}
  \begin{scope}[canvas is xy plane at z=4]
    \fill[orange!80]           ({12-\tokensize/2},{-2-\tokensize/2})
                              rectangle ++({\tokensize},{\tokensize});
  \end{scope}

  % Arrows from grid cells to tokens
  \foreach \i/\j in {0/0,1/0,0/1,1/1}{
    \draw[pink!70,->,thick]
      ({\i*\L/\N+\L/(2*\N)}, {\j*\L/\N+\L/(2*\N)}, 0) -- (T1);
  }
  \foreach \i/\j in {3/5,4/5,3/6,4/6}{
    \draw[blue!60,->,thick]
      ({\i*\L/\N+\L/(2*\N)}, {\j*\L/\N+\L/(2*\N)}, 0) -- (T2);
  }
  \foreach \i/\j in {4/3,5/3,4/4,5/4}{
    \draw[green!70!black,->,thick]
      ({\i*\L/\N+\L/(2*\N)}, {\j*\L/\N+\L/(2*\N)}, 0) -- (T3);
  }
  \foreach \i/\j in {5/0,6/0,5/1,6/1}{
    \draw[orange!80,->,thick]
      ({\i*\L/\N+\L/(2*\N)}, {\j*\L/\N+\L/(2*\N)}, 0) -- (T4);
  }

  % Title
 \node[draw, rounded corners=2pt, fill=black!3, line width=0.9pt,
       font=\sffamily\bfseries, align=center, inner sep=1.6mm]
  (mploss) at (12,1.5,-7.0)
  {Multiple Prompt\\Optimal Transport Loss};
\end{scope}

% Arrow from sim grid into the new 3D panel
\draw[-{Stealth[length=3mm]}, line width=1.2pt, draw=localorange]
  (simgridRight) -- ([xshift=-3mm]otOrigin);

  % --- Frozen badges on ViT layers ---
\path ([xshift=-2mm,yshift=-2mm]vit1.north east) pic {snowbadge};
\path ([xshift=-2mm,yshift=-2mm]vit2.north east) pic {snowbadge};
\path ([xshift=-2mm,yshift=-2mm]vit4.north east) pic {snowbadge};
\path ([xshift=-2mm,yshift=-2mm]vit5.north east) pic {snowbadge};

% --- Frozen badges on V-V attn layers ---
\path ([xshift=-2mm,yshift=-2mm]att1.north east) pic {snowbadge};
\path ([xshift=-2mm,yshift=-2mm]att2.north east) pic {snowbadge};
\path ([xshift=-2mm,yshift=-2mm]att4.north east) pic {snowbadge};
\path ([xshift=-2mm,yshift=-2mm]att5.north east) pic {snowbadge};

% --- Frozen badge on the Text Encoder ---
\path ([xshift=-2mm,yshift=-2mm]textencTop.north east) pic {snowbadge};

\foreach \k in {0,...,3} {
  \coordinate (gNW\k) at (-2.2, -\k*0.9 + 0.75); % NW corner of the green tile
  \path ([xshift=1.0mm,yshift=-2.5mm] gNW\k) pic {firebadge};
}

\foreach \kk in {0,...,3} {%
  \coordinate (lNW\kk) at (2.15, -\kk*0.9 + 0.75); % NW corner of the (commented) local tile
  \path ([xshift=-1.3mm,yshift=-2.5mm] lNW\kk) pic {firebadge};
}

% --- Legend at bottom: badges ---
\coordinate (legendBase) at ([yshift=-6mm]current bounding box.south);

% Frozen badge + label
\path ([xshift=1.5mm,yshift=-3.0mm] projBlock.north west) pic {firebadge};
\path (legendBase) ++(-18mm,0) coordinate (legSnow);
\path (legSnow) pic {snowbadge};
\node[anchor=west, font=\sffamily\scriptsize]
  at ([xshift=2.2mm]legSnow) {Frozen Parameters};

% Fire badge + label
\path (legendBase) ++(18mm,0) coordinate (legFire);
\path (legFire) pic {firebadge};
\node[anchor=west, font=\sffamily\scriptsize]
  at ([xshift=2.2mm]legFire) {Learnable Parameters};

\node[
  draw=blue!60!black,
  rounded corners=1pt,
  line width=0.9pt,
  fit=(vit1)(vit2)(vit4)(vit5)(att1)(att2)(att4)(att5)(dotsT)(dotsB),
  inner sep=1mm
] (imgenc) {};

% label for the box (sits just above the frame)
\node[font=\sffamily\bfseries, anchor=south, yshift=-6.2mm]
  at (imgenc.south) {Image Encoder};

\node[
  draw=black!35,
  rounded corners=2pt,
  inner sep=0pt,
  anchor=east
] (imgin) at ([xshift=-18mm]imgenc.west) {\includegraphics[width=30mm]{metu.jpeg}};

\node[anchor=north, font=\sffamily\scriptsize] at (imgin.south) {Image Input};

% Arrow into the encoder box
\draw[arr] (imgin.east) -- ([xshift=-1mm]imgenc.west);

\end{scope}

\end{tikzpicture}
\end{adjustbox}
\caption{Overview of the SOT-GLP framework. The image encoder uses two parallel streams: a standard CLIP pathway (Q-K attention) for global features and a V-V attention pathway for local patch features. Global prompts generate class-agnostic text embeddings matched to the global [CLS] token, while class-specific local prompts align to Top-$K$ selected patches via Optimal Transport. Both branches contribute to the final classification logits.}
\label{fig:our-arch}
\end{figure*}

We propose SOT-GLP (Sparse Optimal Transport Guided Local-Global Prompt Learning), a framework that preserves CLIP's global alignment while explicitly modeling fine-grained spatial structure. As illustrated in Figure~\ref{fig:our-arch}, our approach employs a dual-branch architecture with four complementary components. The \textbf{global prompt branch} maintains standard CLIP-style image–text matching, preserving robust category-level generalization. The \textbf{local prompt branch} operates on patch tokens extracted via value–value (V–V) attention, enabling the model to capture textures and fine-grained parts. To mitigate background interference, we introduce \textbf{saliency-guided sparsification}, a filtering mechanism that dynamically suppresses non-discriminative clutter to construct a clean semantic support set. Finally, \textbf{Sparse Optimal Transport (OT)} aligns these filtered visual patch features to multiple class-specific local prompts via a balanced set-to-set objective. By enforcing uniform transport marginals, the OT formulation constrains each prompt to receive comparable assignment mass, preventing degenerate solutions where multiple prompts attend to the same dominant patch.

Our contributions include the discovery of a distinct accuracy-robustness trade-off in prompt learning. We demonstrate that while a learnable local projection maximizes few-shot accuracy ($+0.9$\% gain), removing this projection preserves the pre-trained CLIP feature manifold, resulting in \textbf{state-of-the-art Out-of-Distribution detection performance} (23.8 FPR95 / 94.2 AUC), significantly outperforming recent methods.

\section{Related Work}
Our work builds on three research directions: prompt learning, locality-aware feature extraction, and patch-prompt aligment.

\textit{Prompt learning.} Early approaches to prompt learning depended on hand-crafted prompt templates derived directly from human expertise and prior understanding \cite{petroni2019language, poerner2019bert}. CoOp~\cite{zhou2022learning} replaced hand-crafted templates with task-specific soft prompts to improve few-shot performance of CLIP under both class specific and non-class-specific prompt settings. CoCoOp~\cite{zhou2022conditional} extended this with instance-conditioned prompts for better generalization, while MaPLe~\cite{khattak2023maple} jointly learns prompts in both vision and text branches. PromptSRC~\cite{khattak2023self} adds self-regularization to prevent overfitting, and PLOT~\cite{chen2022plot} introduces patch-level matching via OT. State-of-the-art approaches like GalLoP~\cite{lafon2024gallop} combines global and local prompts with multi-scale top-$K$ selection and prompt dropout for diversity. However, GalLoP's independent top-$K$ selection per prompt can lead to overlapping support and implicit competition. Unlike GalLoP, which performs independent top-K selection per prompt and permits overlapping support, we perform a single shared top-K selection and explicitly allocate patches to prompts via balanced optimal transport, enforcing non-overlapping specialization by construction.

\textit{Enhancing locality in the visual encoder.}
Standard CLIP attention produces globally-pooled features that may lose fine-grained spatial details. CLIP Surgery~\cite{li2025closer} replaces query-key with value-value (V-V) attention to strengthen patch-level interactions, yielding more discriminative local representations. Departing from standard prompt learning approaches that rely on CLIP's global-biased self-attention, we are the first to repurpose Value-Value (V-V) attention as a dedicated local feature stream, explicitly amplifying patch-to-patch correlations to drive fine-grained prompt alignment.

\textit{Patch-prompt alignment.}
CoOp-based methods ~\cite{zhou2022learning} match global image features to learned text prompts, but this whole-image objective obscures region-level details. PLOT~\cite{chen2022plot} addresses this with dense optimal transport (OT) over all patch-prompt pairs, yielding fine-grained alignment but at high computational cost and with background interference. GalLoP~\cite{lafon2024gallop} restricts alignment to top-$K$ patches per prompt, selected independently for each prompt, and employs multi-scale training where different prompts use different $K$ values to capture parts at varying scales. While efficient, independent selection allows prompts to attend to overlapping patches, reducing diversity. We combine the efficiency of sparse selection with the balanced assignment guarantees of OT: we first identify a shared top-$K$ set of salient patches, then solve entropically regularized OT~\cite{cuturi2013sinkhorn} to allocate them among class-specific local prompts. This formulation (i)~enforces balanced coverage, preventing few prompts from dominating; (ii)~yields consistent gradients by softly partitioning patches among prompts; and (iii)~remains fully differentiable via Sinkhorn iterations, improving convergence while preserving diversity.

\section{Method}
\label{sec:approach}
We introduce dual prompt branches: global prompts maintain category-level alignment via whole-image features, while local prompts capture fine-grained details by aligning salient patches—extracted through V-V attention—to class-specific embeddings via Optimal Transport.

\subsection{Global and Local Prompt Parameterization}

Let $\mathcal{C}$ denote the set of classes. We learn two sets of prompts. First, we define a set of \textbf{global prompts} $\mathcal{P}^{g} = \{\mathbf{p}^{g}_m\}_{m=1}^{N_g}$, where each $\mathbf{p}^{g}_m \in \mathbb{R}^{M \times d}$ is a learnable token sequence shared across all classes. Global prompts provide a shared contextual foundation that regularizes the text branch and reduces overfitting in few-shot regimes, maintaining the stability benefits observed in CoOp~\cite{zhou2022learning}. 

Second, we learn \textbf{local prompts} $\mathcal{P}^{\ell}_c = \{\mathbf{p}^{\ell}_{c,j}\}_{j=1}^{N_\ell}$ specific to each class $c \in \mathcal{C}$, where $N_\ell$ is the number of local prompts per class. Local prompts enable class-specific specialization, capturing discriminative attributes such as textures, part configurations, or fine-grained details that vary across categories.

We initialize global prompt parameters from the tokenized template $\tau_{\text{init}} = \text{``a photo of a''}$, setting the prompt length $M = |\tau_{\text{init}}|$ (number of tokens). Let $\tau(c)$ denote the tokenized class name, and $f_T(\cdot)$ the CLIP text encoder. We generate the text embeddings as follows:

\begin{equation}
\begin{split}
T^{g}_{c,m} &= f_T([\texttt{[SOS]};\, \mathbf{p}^{g}_m;\, \tau(c);\, \texttt{[EOT]}]), \\
&\quad m = 1,\ldots,N_g \\
T^{\ell}_{c,j} &= f_T([\texttt{[SOS]};\, \mathbf{p}^{\ell}_{c,j};\, \tau(c);\, \texttt{[EOT]}]), \\
&\quad j = 1,\ldots,N_\ell
\end{split}
\end{equation}

We collect the local embeddings into a set $\mathcal{L}_c = \{T^{\ell}_{c,j}\}_{j=1}^{N_\ell}$ for alignment with image patches.

\subsection{Locality-Aware Feature Extraction}
\label{sec:vv}

To extract both global and local features, we employ a dual-stream vision encoder. The original CLIP stream preserves global semantics via standard query-key (Q-K) self-attention, while a parallel value-value (V-V) stream enhances local discriminability. Unlike Q-K attention, which computes similarity between query and key projections, V-V attention directly correlates value representations~\cite{li2025closer}, strengthening patch-to-patch interactions and producing locality-aware features with reduced background interference. This approach has proven effective for such as anomaly detection~\cite{li2024promptad} while preserving global performance. We extract $Z_{\text{global}}$ from the original stream's [CLS] token and $Z_{\text{local}}$ from the V-V stream's patch tokens, enabling complementary global-local alignment.

Let the token sequence at layer $l$ be $Z_{l} = [\,t^{l}_{\text{cls}};\,t^{l}_{1};\ldots;t^{l}_{T}\,] \in \mathbb{R}^{(T+1)\times d}$, where $t^{l}_{\text{cls}}$ is the class token and $\{t^{l}_p\}_{p=1}^T$ are patch tokens. We process this sequence through $L$ transformer blocks via two parallel streams, both initialized from the same input: $Z^{\text{ori}}_{0} = Z^{\text{vv}}_{0} = Z_{0}$.

\textit{Original CLIP stream.}
We process tokens through standard transformer blocks with Q-K attention~\cite{radford2021learning}, extracting the global embedding from the final [CLS] token: $Z_{\text{global}} = (Z^{\text{ori}}_{L})_{0} \in \mathbb{R}^{d}$.

\textit{V-V attention stream.}
In parallel, we compute V-V attention by directly correlating value representations:
\begin{equation}
\begin{split}
A^{\text{vv}}_l &= \text{softmax}\!\left(\frac{V_l V_l^\top}{\sqrt{d}}\right), \quad
Y^{\text{vv}}_l = A^{\text{vv}}_l V_l, \\
Z^{\text{vv}}_l &= Z^{\text{vv}}_{l-1} + \text{Proj}^{v}_l(Y^{\text{vv}}_l)
\label{eq:vv}
\end{split}
\end{equation}
The final local features are obtained by extracting the non-\texttt{[CLS]} tokens and passing them through a learnable linear projection $W_{\text{proj}} \in \mathbb{R}^{d \times d}$:
\begin{equation}
Z_{\text{local}} \;=\; \big(Z^{\text{vv}}_{L}\big)_{1:T} \, W_{\text{local-proj}}
\label{eq:zlocal}
\end{equation}

\subsection{Global Image-Text Alignment}
\label{sec:global}

The global branch preserves CLIP's standard contrastive objective, matching whole-image representations to learned text prompts.
For a minibatch $\{(I_i, c_i)\}_{i=1}^{B}$, where $I_i$ is an image and $c_i \in \mathcal{C}$ its ground-truth class, $Z^{(i)}_{\text{global}}$ is the corresponding global image embedding and $\{T^{g}_{c,m}\}_{m=1}^{N_g} \subset \mathbb{R}^{d}$ the $N_g$ global text prompt embeddings for class $c$. We compute the similarity score between image $i$ and class $c$ by averaging over all $N_g$ global prompts with temperature scaling $\tau > 0$:
\begin{equation}
s^{\text{global}}_{i,c} = \frac{1}{N_g}\sum_{m=1}^{N_g} \frac{(Z^{(i)}_{\text{global}})^{\!\top} T^{g}_{c,m}}{\tau}
\label{eq:global-scores}
\end{equation}
This induces a categorical distribution over classes via softmax:
\begin{equation}
p_{\theta}(c \mid I_i) = \frac{\exp(s^{\text{global}}_{i,c})}{\sum_{c' \in \mathcal{C}} \exp(s^{\text{global}}_{i,c'})}
\label{eq:global-prob}
\end{equation}
and the global contrastive loss is the standard cross-entropy:
\begin{equation}
\mathcal{L}_{\text{global}} = -\frac{1}{B}\sum_{i=1}^{B} \log p_{\theta}(c_i \mid I_i)
\label{eq:global-loss}
\end{equation}
This objective encourages the learned global prompts to maximize similarity with images from their corresponding class while remaining discriminative across all classes. Additionally, to prevent redundancy and encourage diversity among the learned global prompts, we employ global prompt dropout during training.

\subsection{Sparse Optimal Transport Alignment}
\label{sec:local}

While the global branch captures category-level semantics, fine-grained discrimination often depends on local visual details—textures, part configurations, or spatial patterns. However, dense alignment methods \cite{chen2022plot} suffer from \textit{background dominance}, where mass is distributed to irrelevant regions. We address background dominance and prompt redundancy by reformulating local prompt alignment as a two-stage process: (i) shared saliency-guided sparsification to define a common support set, and (ii) balanced entropic optimal transport to partition this support across prompts.

\textit{Saliency-Guided Sparsification.}
Consider an image $I_i$ with patch tokens $\{z^{\ell}_{i,p}\}_{p=1}^{P}$ from the V-V stream. To evaluate alignment with a specific class $c \in \mathcal{C}$, let $\mathcal{L}_{c} = \{t^{\ell}_{c,j}\}_{j=1}^{N_\ell}$ denote the set of $N_\ell$ class-specific local prompt embeddings.
Directly aligning all patches incorporates noise. Instead, we define a \textbf{saliency map} for the image relative to class $c$ by computing the mean similarity of each patch against the local prompt ensemble:
\begin{equation}
\sigma^{(i)}_{p,c} = \frac{1}{N_\ell}\sum_{j=1}^{N_\ell} (z^{\ell}_{i,p})^{\!\top} t^{\ell}_{c,j}
\label{eq:patch-scores}
\end{equation}
We then apply a sparsification operator $\mathcal{T}_{\text{sparse}}$ which retains only the subset of visual primitives that maximize this semantic density. Formally, we define the \textbf{sparse support set} $\mathcal{S}^{(i)}_c$:
\begin{equation}
\begin{split}
\mathcal{S}^{(i)}_c &= \mathcal{T}_{\text{sparse}}\left(\{z^{\ell}_{i,p}\}_{p=1}^{P} \mid \sigma^{(i)}_{p,c}\right) \\
&= \{z^{\ell}_{i,p} : \text{rank}(\sigma^{(i)}_{p,c}) \leq K\}
\end{split}
\label{eq:topk-selection}
\end{equation}
This operation filters non-discriminative background clutter, ensuring that the subsequent transport problem is well-conditioned and grounded solely on salient foreground regions.

\textit{Balanced Entropic Transport.}
Given the sparse support set $\mathcal{S}^{(i)}_c = \{s_u\}_{u=1}^{K}$ and local prompts $\mathcal{L}_{c} = \{t_v\}_{v=1}^{N_\ell}$, we formulate the alignment as an Optimal Transport problem. We compute pairwise similarities $\text{sim}_{uv} = s_u^{\top} t_v$ and define the transport cost $C_{uv} = 1 - \text{sim}_{uv}$. To prevent \textit{prompt collapse} (where all prompts attend to the same single "best" patch), we enforce a balanced transport plan $\mathbf{T} \in \mathbb{R}^{K \times N_\ell}$ with uniform marginals $\mathbf{a} = \frac{1}{K}\mathbf{1}_K$ and $\mathbf{b} = \frac{1}{N_\ell}\mathbf{1}_{N_\ell}$:
\begin{equation}
\begin{split}
\mathbf{T} &= \text{Sinkhorn}(C, \mathbf{a}, \mathbf{b}), \\
\text{s.t.} \quad & \mathbf{T}\mathbf{1} = \mathbf{a}, \; \mathbf{T}^{\top}\mathbf{1} = \mathbf{b}
\end{split}
\label{eq:sinkhorn}
\end{equation}
The balanced marginal constraints prevent any single prompt from dominating all patches (prompt collapse) and encourage diversified assignments where different prompts specialize on different visual parts. The transport-weighted local similarity score for class $c$ is:
\begin{equation}
\phi^{(i)}_{c} = \sum_{u=1}^{K}\sum_{v=1}^{N_\ell} T_{uv} \cdot \text{sim}_{uv}
\label{eq:local-score}
\end{equation}
We define the temperature-scaled local logit $s^{\text{local}}_{i,c}$ and obtain class probabilities via softmax:
\begin{equation}
\begin{split}
s^{\text{local}}_{i,c} &= \frac{\phi^{(i)}_{c}}{\tau}, \\
p^{\text{loc}}_{\theta}(c \mid I_i) &= 
\frac{\exp(s^{\text{local}}_{i,c})}{\sum_{c' \in \mathcal{C}} \exp(s^{\text{local}}_{i,c'})}
\end{split}
\label{eq:local-prob}
\end{equation}
The local loss $\mathcal{L}_{\text{local}}$ is computed as the cross-entropy over $p^{\text{loc}}_{\theta}$. This formulation combines the computational efficiency of sparse Top-$K$ selection with the balanced assignment guarantees of OT, yielding fine-grained patch-prompt correspondences that complement the global branch.

\subsection{Training Objective and Inference}
\label{sec:objective}

The complete training objective combines the global and local losses:
$
\mathcal{L} = \mathcal{L}_{\text{global}} + \lambda \mathcal{L}_{\text{local}}
$
where $\lambda > 0$ determines the strength of the local supervision relative to the global baseline. These objectives are complementary: the global term provides stable, transferable representations that prevent overfitting, while the local term increases sensitivity to spatially localized attributes. During training, both branches are optimized jointly with respect to all learnable prompt parameters $\{\mathbf{P}^g, \{\mathbf{P}^{\ell}_c\}_{c \in \mathcal{C}}\}$ and local projection while keeping the vision and text encoders frozen.

At test time, for each class, we compute both global score (Eq.~\ref{eq:global-scores}) and local similarity scores via Top-$K$ selection (Eq.~\ref{eq:topk-selection}) and OT matching (Eq.~\ref{eq:local-score}). The final classification score combines both branches: $s_{i,c}^{\text{final}} = s_{i,c}^{\text{global}} + \lambda s^{\text{local}}_{i,c}
\label{eq:final-score}
$
followed by softmax normalization to obtain class probabilities. This weighted additive fusion allows the global branch to provide robust baseline predictions while the local branch refines discrimination on fine-grained categories. For OOD detection, we follow the GL-MCM protocol~\cite{miyai2025gl} and compute OOD scores by combining global CLIP similarity with locality-enhanced similarity from the V-V branch, enabling the model to detect distributional shifts based on both global semantics and local feature consistency.

\section{Experimental Results}

We evaluate on two complementary objectives: (1)~few-shot classification on 11 standard benchmarks, reporting Top-1 accuracy with $\{1, 2, 4, 8, 16\}$ shots per class, and (2)~out-of-distribution (OOD) detection on ImageNet-centric test suites, where the model is expected to distinguish in-distribution ImageNet samples from OOD examples. For OOD detection, we report FPR95 (false positive rate at 95\% true positive rate) and AUROC (area under the ROC curve), following the GL-MCM protocol~\cite{miyai2025gl}.

\textit{Datasets.} Our evaluation spans diverse visual domains: \textbf{generic objects} (ImageNet~\cite{deng2009imagenet}, Caltech101~\cite{fei2004learning}); \textbf{fine-grained categories} (FGVC Aircraft~\cite{maji2013fine}, Stanford Cars~\cite{krause20133d}, Oxford Pets~\cite{parkhi2012cats}) where subtle part-level differences dominate; \textbf{textures} (DTD~\cite{cimpoi2014describing}) emphasizing local patterns; \textbf{scenes} complex spatial layouts; \textbf{remote sensing} (EuroSAT~\cite{helber2019eurosat}); \textbf{food} (Food101~\cite{bossard2014food}); \textbf{flowers} (Flowers102~\cite{nilsback2008automated}); and \textbf{actions} (UCF101~\cite{soomro2012ucf101}) using RGB frames.

\textit{Implementation details.} All experiments use CLIP ViT-B/16 with frozen vision and text encoders; only prompt parameters are learned. We learn 4 global prompts (shared across classes) and 4 local prompts per class, with Top-$K = 10$ patches selected for OT matching. Training uses SGD with learning rate 0.05, momentum 0.9, weight decay 0.01, and cosine annealing over 50 epochs with 5-epoch warmup. We train on 4 NVIDIA RTX 3090 GPUs with mixed-precision and data-parallel execution, maintaining an effective batch size of 32. We follow the standard fixed-seed, fixed-split protocol and report results averaged over 3 random initializations for the 16-shot setting. Hyperparameters (learning rate, number of prompts, Top-$K$) are kept constant across datasets to demonstrate generalizability rather than per-dataset tuning.

\textit{Baselines.} We compare against representative CLIP adaptation methods: CLIP (zero-shot with hand-crafted prompts)~\cite{radford2021learning}; Linear Probe (linear classifier on frozen features); and prompt learning methods CoOp~\cite{zhou2022learning} (learnable prompts), CoCoOp~\cite{zhou2022conditional} (image-conditioned prompts), MaPLe~\cite{khattak2023maple} (vision-language co-prompting), PLOT~\cite{chen2022plot} (dense OT alignment), PromptSRC~\cite{khattak2023self} (self-regularized prompts), ProDA~\cite{lu2022prompt} (prompt distribution learning), LoCoOp~\cite{miyai2023locoop} (OOD-aware prompts), and GalLoP~\cite{lafon2024gallop} (global-local prompts).

\begin{table}[t]
\centering
\caption{Few-shot classification accuracy (\%) on 11 benchmarks with 16 shots per class using CLIP ViT-B/16.}
\label{tab:16shot-vitb16-accuracy-transposed}

\sffamily 
\footnotesize

\setlength{\tabcolsep}{1.6pt} 
\renewcommand{\arraystretch}{1.25} 
\rowcolors{2}{gray!10}{white}

\newcolumntype{Y}{>{\centering\arraybackslash}X}

\newcommand{\rhead}[2]{%
  \rotatebox{90}{%
    \textbf{#1}% 
    \ifx&#2&% 
    \else
      ~{\color{blue!80!black}#2}%
    \fi
  }%
}

\begin{tabularx}{\linewidth}{l *{11}{Y}} 
\toprule

\textbf{Dataset} &
\rhead{CLIP}{\cite{radford2021learning}} &
\rhead{Linear Probe}{} &   % <--- Fixed: Linear Probe combined
\rhead{CoOp}{\cite{zhou2022learning}} &
\rhead{CoCoOp}{\cite{zhou2022conditional}} &
\rhead{MaPLe}{\cite{khattak2023maple}} &
\rhead{PLOT}{\cite{chen2022plot}} &
\rhead{PromptSRC}{\cite{khattak2023self}} &
\rhead{LoCoOp}{\cite{miyai2023locoop}} &
\rhead{ProDA}{\cite{lu2022prompt}} &
\rhead{GalLoP}{\cite{lafon2024gallop}} &
\rhead{SOT-GLP}{} \\
\midrule

% Data Rows
ImageNet   & 66.7 & 67.3 & 71.7 & 71.0 & 72.3 & 72.6 & 73.2 & 71.5 & 71.9 & \underline{75.1} & \textbf{75.5} \\
Caltech101 & 92.2 & 95.4 & 95.6 & 95.2 & 96.0 & 96.0 & 96.1 & 94.9 & 95.5 & \underline{96.7} & \textbf{97.4} \\
OxfordPets & 88.4 & 85.3 & 91.9 & 93.3 & 92.8 & 93.6 & 93.7 & 92.4 & 93.5 & \underline{94.1} & \textbf{94.8} \\
Cars       & 65.5 & 80.4 & 83.1 & 71.6 & 83.6 & 84.6 & 85.8 & 79.8 & 79.8 & \textbf{89.2} & \textbf{89.2} \\
Flowers102 & 70.7 & 97.4 & 97.1 & 87.8 & 97.0 & 97.6 & 97.6 & 96.3 & 96.8 & \underline{98.8} & \textbf{99.2} \\
Food101    & 84.8 & 82.9 & 84.2 & \underline{87.2} & 85.3 & 87.1 & 86.5 & 84.7 & 86.8 & 86.5 & \textbf{87.8} \\
Aircraft   & 24.8 & 45.4 & 43.4 & 31.2 & 48.4 & 46.7 & 50.8 & 40.7 & 40.2 & \textbf{58.3} & \underline{57.6} \\
SUN397     & 62.3 & 73.3 & 74.7 & 72.2 & 75.5 & 76.0 & \underline{77.2} & 74.2 & 75.7 & \underline{77.2} & \textbf{78.2} \\
DTD        & 44.1 & 70.0 & 69.9 & 63.0 & 71.3 & 71.4 & 72.7 & 69.5 & 70.9 & \underline{75.5} & \textbf{77.1} \\
EuroSAT    & 48.3 & 87.2 & 84.9 & 73.3 & \underline{92.3} & 92.0 & \textbf{92.4} & 86.1 & 85.1 & 90.1 & 91.7 \\
UCF101     & 64.7 & 82.1 & 82.2 & 78.1 & 85.0 & 85.3 & 86.5 & 81.6 & 83.3 & \underline{86.9} & \textbf{87.5} \\
\midrule
\textbf{Average} & 75.7 & 78.8 & 79.9 & 74.9 & 81.8 & 82.1 & 82.9 & 79.2 & 80.0 & \underline{84.4} & \textbf{85.1} \\
\bottomrule
\multicolumn{12}{l}{\textit{Best per row in \textbf{bold}, second best \underline{underlined}.}} \\
\end{tabularx}
\end{table}

\textit{Few-Shot Classification Results.}
Table~\ref{tab:16shot-vitb16-accuracy-transposed} presents 16-shot classification accuracy on 11 benchmarks. SOT-GLP achieves the best average accuracy (85.1\%), outperforming all prompt-learning baselines and sets state-of-the-art results on 9 out of 11 datasets (ImageNet, Caltech101, OxfordPets, Cars, Flowers102, Food101, SUN397, DTD, UCF101) and remains competitive on the remaining two (Aircraft, EuroSAT). For detailed performance breakdowns across all shot settings ($K \in \{1, 2, 4, 8, 16\}$), please refer to Table~\ref{tab:fewshot_compact_vertical}.

Gains are most pronounced on datasets where local features complement global semantics: texture recognition (DTD), fine-grained flowers (Flowers102), and action recognition (UCF101). Improvements on generic object recognition (ImageNet) and scene classification (SUN397), demonstrate that local-global fusion benefits both coarse and fine-grained tasks.
On Aircraft and EuroSAT, SOT-GLP trails slightly. Aircraft's extreme fine-grained nature (distinguishing aircraft models by subtle fuselage or wing details) makes it highly sensitive to whether selected patches capture discriminative parts; CLIP's pretraining may not provide ideal patch representations for such specialized domains. On EuroSAT, our local branch actually outperforms the global branch individually, but additive fusion pulls the combined score below the local-only result, suggesting that weighted fusion or gating mechanisms could improve performance. Despite these cases, SOT-GLP achieves the highest average across all datasets, confirming that combining global alignment with OT-guided local matching provides robust, generalizable improvements.

\begin{figure*}[t]
\centering

% ---------------- GLOBAL COLORS / MARKERS ----------------
\definecolor{cclipmcm}{RGB}{150,150,150}
\definecolor{cclipgl}{RGB}{120,120,120}
\definecolor{ccoopmcm}{RGB}{227,119,194}
\definecolor{ccoopgl}{RGB}{199,74,162}
\definecolor{cplot}{RGB}{31,119,180}
\definecolor{clocoopm}{RGB}{44,160,44}
\definecolor{clocoopgl}{RGB}{30,110,30}
\definecolor{cpromptsrc}{RGB}{255,127,14}
\definecolor{clsncoop}{RGB}{148,103,189}
\definecolor{clsncocoop}{RGB}{107,70,154}
\definecolor{cgallop}{RGB}{214,39,40}
\definecolor{cours}{RGB}{0,114,178}     % darker blue for ours
\definecolor{coursnol}{RGB}{230,159,0}  % strong orange for no-local

\tikzset{
  mclipmcm/.style={cclipmcm,mark=o,mark size=1.7pt,line width=0.3pt},
  mclipgl/.style={cclipgl,mark=*,mark size=1.5pt,line width=0.3pt},
  mcoopmcm/.style={ccoopmcm,mark=triangle*,mark size=1.7pt,line width=0.3pt},
  mcoopgl/.style={ccoopgl,mark=triangle,mark size=1.8pt,line width=0.3pt},
  mplot/.style={cplot,mark=square*,mark size=1.6pt,line width=0.3pt},
  mlocoopm/.style={clocoopm,mark=diamond*,mark size=1.7pt,line width=0.3pt},
  mlocoopgl/.style={clocoopgl,mark=diamond,mark size=1.8pt,line width=0.3pt},
  mprompt/.style={cpromptsrc,mark=asterisk,mark size=2.0pt,line width=0.35pt},
  mlsncoop/.style={clsncoop,mark=oplus*,mark size=1.8pt,line width=0.3pt},
  mlsncocoop/.style={clsncocoop,mark=oplus,mark size=1.9pt,line width=0.3pt},
  mgallop/.style={cgallop,mark=otimes*,mark size=1.4pt,line width=0.35pt},
  mours/.style={cours,mark=star,mark size=4.0pt,line width=0.75pt,fill=white},
  moursnol/.style={coursnol,mark=star,mark size=4.0pt,line width=0.75pt},
}

% ====== LEFT ======
\begin{minipage}[t]{0.36\linewidth}
\centering
\begin{tikzpicture}[scale=0.78, x=0.58cm,y=0.28cm]
  \scriptsize
  \def\xmin{66}\def\xmax{76}
  \def\ymin{23}\def\ymax{44}

  \fill[gray!3] (\xmin,\ymin) rectangle (\xmax,\ymax);
  \foreach \x in {66,68,70,72,74,76} \draw[very thin,gray!30] (\x,\ymin) -- (\x,\ymax);
  \foreach \y in {24,28,32,36,40,44} \draw[very thin,gray!30] (\xmin,\y) -- (\xmax,\y);

  \draw[semithick] (\xmin,\ymin) -- (\xmax,\ymin);
  \draw[semithick] (\xmin,\ymin) -- (\xmin,\ymax);

  \node[rotate=90,anchor=center] at ([xshift=-26pt]\xmin,{(\ymin+\ymax)/2}) {Avg. FPR95 (↓)};
  \node[below=10pt] at ({(\xmin+\xmax)/2},\ymin) {ImageNet Top-1 (\%)};

  \foreach \x/\lab in {66/66,68/68,70/70,72/72,74/74,76/76} {
    \draw[semithick] (\x,\ymin) -- (\x,\ymin-0.35);
    \node[below=2.5pt] at (\x,\ymin) {\lab};
  }
  \foreach \y/\lab in {24/24,28/28,32/32,36/36,40/40,44/44} {
    \draw[semithick] (\xmin,\y) -- (\xmin-0.25,\y);
    \node[left=2.5pt] at (\xmin,\y) {\lab};
  }

  \draw[mclipmcm] plot coordinates {(66.7,42.8)};
  \draw[mclipgl] plot coordinates {(66.7,35.5)};
  \draw[mcoopmcm] plot coordinates {(71.7,36.8)};
  \draw[mcoopgl] plot coordinates {(71.7,30.7)};
  \draw[mplot] plot coordinates {(72.6,31.8)};
  \draw[mlocoopm] plot coordinates {(71.5,34.0)};
  \draw[mlocoopgl] plot coordinates {(71.5,28.7)};
  \draw[mprompt] plot coordinates {(73.2,33.7)};
  \draw[mlsncoop] plot coordinates {(72.9,32.0)};
  \draw[mlsncocoop] plot coordinates {(72.9,30.2)};
  \draw[mgallop] plot coordinates {(75.1,27.3)};

  % ==== Our points: draw last so they are on top ====
  \draw[mours] plot coordinates {(75.5,28.1)};
  \draw[moursnol] plot coordinates {(75.4,23.8)};
\end{tikzpicture}
\end{minipage}
\hspace{1mm}
% ====== RIGHT ======
\begin{minipage}[t]{0.36\linewidth}
\centering
\begin{tikzpicture}[scale=0.78, x=0.58cm,y=1.176cm]
  \scriptsize
  \def\xmin{66}\def\xmax{76}
  \def\ymin{90}\def\ymax{95}

  \fill[gray!3] (\xmin,\ymin) rectangle (\xmax,\ymax);
  \foreach \x in {66,68,70,72,74,76} \draw[very thin,gray!30] (\x,\ymin) -- (\x,\ymax);
  \foreach \y in {91,93,95} \draw[very thin,gray!30] (\xmin,\y) -- (\xmax,\y);

  \draw[semithick] (\xmin,\ymin) -- (\xmax,\ymin);
  \draw[semithick] (\xmin,\ymin) -- (\xmin,\ymax);

  \node[rotate=90,anchor=center] at ([xshift=-26pt]\xmin,{(\ymin+\ymax)/2}) {Avg. AUC (↑)};
  \node[below=10pt] at ({(\xmin+\xmax)/2},\ymin) {ImageNet Top-1 (\%)};

  \foreach \x/\lab in {66/66,68/68,70/70,72/72,74/74,76/76} {
    \draw[semithick] (\x,\ymin) -- (\x,\ymin-0.2);
    \node[below=2.5pt] at (\x,\ymin) {\lab};
  }
  \foreach \y/\lab in {91/91,93/93,95/95} {
    \draw[semithick] (\xmin,\y) -- (\xmin-0.22,\y);
    \node[left=2.5pt] at (\xmin,\y) {\lab};
  }

  \draw[mclipmcm] plot coordinates {(66.7,90.8)};
  \draw[mclipgl] plot coordinates {(66.7,90.8)};
  \draw[mcoopmcm] plot coordinates {(71.7,91.9)};
  \draw[mcoopgl] plot coordinates {(71.7,91.8)};
  \draw[mplot] plot coordinates {(72.6,92.7)};
  \draw[mlocoopm] plot coordinates {(71.5,92.7)};
  \draw[mlocoopgl] plot coordinates {(71.5,93.5)};
  \draw[mprompt] plot coordinates {(73.2,92.4)};
  \draw[mlsncoop] plot coordinates {(72.9,92.3)};
  \draw[mlsncocoop] plot coordinates {(72.9,93.0)};
  \draw[mgallop] plot coordinates {(75.1,93.2)};

  % ours last
  \draw[mours] plot coordinates {(75.5,93.2)};
  \draw[moursnol] plot coordinates {(75.4,94.2)};
\end{tikzpicture}
\end{minipage}

% ====== LEGEND BELOW ======
\vspace{4pt}
\begin{tikzpicture}
  \scriptsize
  % re-declare styles for legend
  \tikzset{
    mclipmcm/.style={black,mark=o,mark size=1.7pt,line width=0.3pt},
    mclipgl/.style={gray!60,mark=*,mark size=1.5pt,line width=0.3pt},
    mcoopmcm/.style={magenta!70!black,mark=triangle*,mark size=1.7pt,line width=0.3pt},
    mcoopgl/.style={magenta!40!black,mark=triangle,mark size=1.8pt,line width=0.3pt},
    mplot/.style={blue!70!black,mark=square*,mark size=1.6pt,line width=0.3pt},
    mlocoopm/.style={green!60!black,mark=diamond*,mark size=1.7pt,line width=0.3pt},
    mlocoopgl/.style={green!30!black,mark=diamond,mark size=1.8pt,line width=0.3pt},
    mprompt/.style={orange!80!black,mark=asterisk,mark size=2.0pt,line width=0.35pt},
    mlsncoop/.style={purple!70!black,mark=oplus*,mark size=1.8pt,line width=0.3pt},
    mlsncocoop/.style={purple!40!black,mark=oplus,mark size=1.9pt,line width=0.3pt},
    mgallop/.style={red!70!black,mark=otimes*,mark size=1.9pt,line width=0.35pt},
    mours/.style={cours,mark=star,mark size=3.0pt,line width=0.7pt,fill=white},
    moursnol/.style={coursnol,mark=star,mark size=2.7pt,line width=0.7pt},
  }

  \node[draw=black!15,rounded corners=2pt,fill=white,inner sep=4pt] {
    \begin{tabular}{@{}llll@{}}
      \tikz{\draw[mclipmcm] plot coordinates {(0,0)};}   & CLIP--MCM
      & \tikz{\draw[mcoopmcm] plot coordinates {(0,0)};} & CoOp--MCM \\
      \tikz{\draw[mclipgl] plot coordinates {(0,0)};}    & CLIP--GL
      & \tikz{\draw[mcoopgl] plot coordinates {(0,0)};}  & CoOp--GL \\
      \tikz{\draw[mplot] plot coordinates {(0,0)};}      & PLOT
      & \tikz{\draw[mlocoopm] plot coordinates {(0,0)};} & LoCoOp--M \\
      \tikz{\draw[mlocoopgl] plot coordinates {(0,0)};}  & LoCoOp--GL
      & \tikz{\draw[mprompt] plot coordinates {(0,0)};}  & PromptSRC \\
      \tikz{\draw[mlsncoop] plot coordinates {(0,0)};}   & LSN+CoOp
      & \tikz{\draw[mlsncocoop] plot coordinates {(0,0)};} & LSN+CoCoOp \\
      \tikz{\draw[mgallop] plot coordinates {(0,0)};}    & GalLoP
      & \tikz{\draw[mours] plot coordinates {(0,0)};}    & \textbf{SOT-GLP} \\
      \tikz{\draw[moursnol] plot coordinates {(0,0)};}   & \textbf{SOT-GLP (No Local Proj) } &
    \end{tabular}
  };
\end{tikzpicture}

\caption[Avg. FPR95 and Avg. AUC vs. ImageNet accuracy trade-offs.]{Left: Avg.\ FPR95 vs.\ ImageNet accuracy. Right: Avg.\ AUC vs.\ ImageNet accuracy. Our method (blue star) is in the high-accuracy / low-FPR95 corner (75.5, 28.1), achieving very similar FPR95 and AUC to GalLoP but at higher ImageNet accuracy, while the variant without local projection (orange star) further lowers FPR95 to 23.8 and increases AUC score to 94.2 with only a small drop in accuracy. No local projection variant shows state of art OOD performance compared to the other prompt-learning baselines, with
\textbf{FPR95} gains ranging from \(\downarrow 4.9\) (vs.\ LoCoOp--GL, \(28.7 \to 23.8\)) to \(\downarrow 3.5\) (vs.\ Gallop, \(27.3 \to 23.8\)), and \textbf{AUC} gains of \(\uparrow 0.7\) (vs.\ LoCoOp (GL-MCM), \(93.5 \to 94.2\)) up to \(\uparrow 1.0\) (vs.\ Gallop, \(93.2 \to 94.2\)).}
\label{fig:ood-fixed-caption}
\end{figure*}
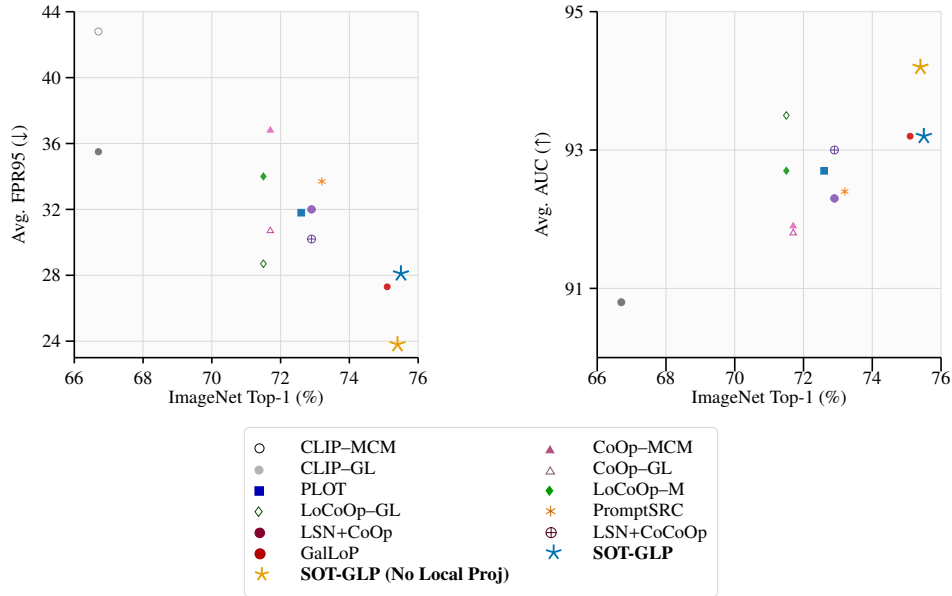

\textit{Out-of-Distribution Detection Results.}
Table~\ref{tab:ood-fullnames} reports OOD detection performance on ImageNet-based benchmarks, where the model must distinguish in-distribution ImageNet samples from out-of-distribution examples (iNaturalist~\cite{vanhorn2018inaturalist}, SUN~\cite{xiao2010sun}, Places~\cite{zhou2017places}, and Textures~\cite{cimpoi2014describing}). SOT-GLP achieves 28.1 FPR95 and 93.2 AUC, competitive with the best prior methods. Notably, the variant without learnable local projection (SOT-GLP w/o proj.) has \textbf{23.8 FPR95} and \textbf{94.2 AUC}, outperforming all prompt-learning baselines while maintaining 75.4\% ImageNet accuracy (only 0.1\% below the full model). This demonstrates a favorable accuracy-robustness trade-off as illustrated in Figure~\ref{fig:ood-fixed-caption}: removing task-adaptive projection in the local branch preserves more of CLIP's pretrained representation space, yielding better-calibrated confidence estimates under distribution shift. The full model prioritizes in-distribution accuracy, while the projection-free variant prioritizes OOD detection—providing practitioners with a simple configuration choice based on deployment requirements.

\begin{figure*}[t]
    \centering
   
    \begin{subfigure}{\textwidth}
        \centering
        \includegraphics[width=0.9\linewidth]{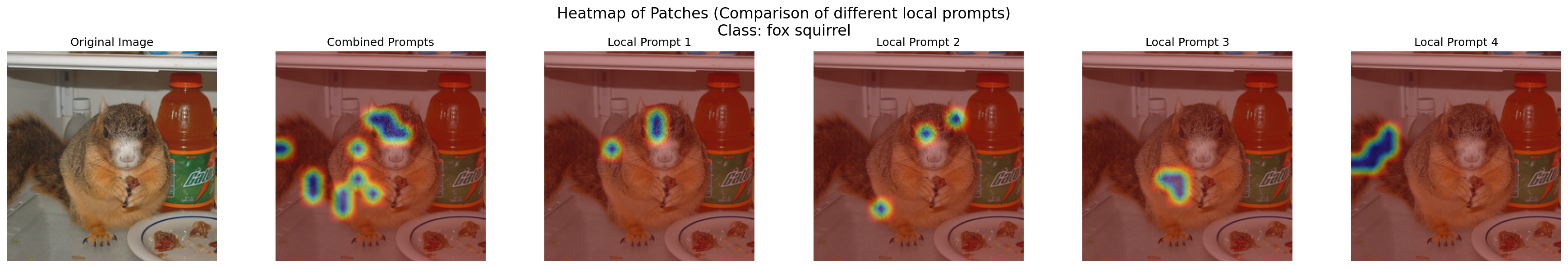}
    \end{subfigure}
    \vspace{-2mm} 

    \begin{subfigure}{\textwidth}
        \centering
        \includegraphics[width=0.9\linewidth]{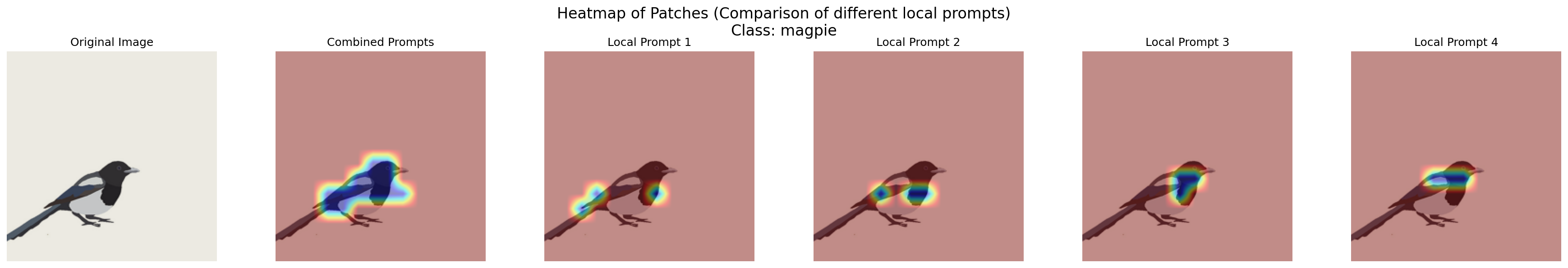}
    \end{subfigure}

    \begin{subfigure}{\textwidth}
        \centering
        \includegraphics[width=0.9\linewidth]{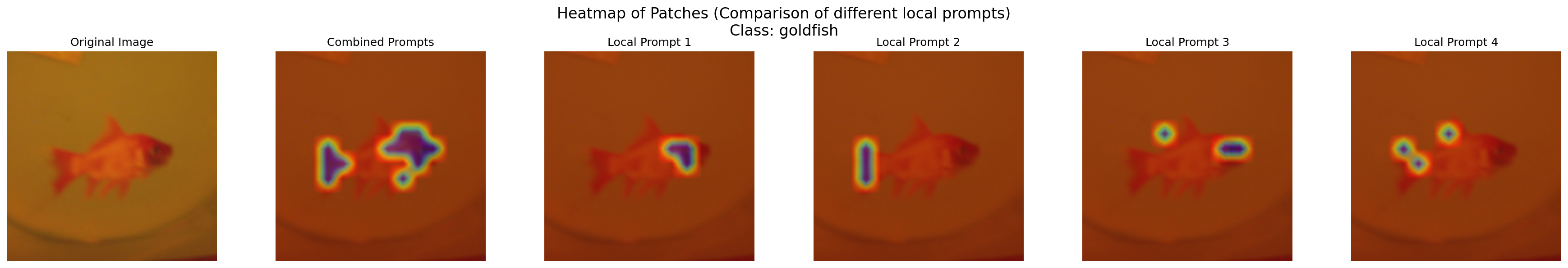}
    \end{subfigure}
    
    \caption{\textbf{Learned Local Prompt Specialization.} Patch–prompt similarity maps for two example classes. For each individual prompt, the top-3 patches are shown. The "combined prompts" column shows similarity between the mean of the local prompts and patch features, computed over top-10 patches. Optimal Transport constraint promotes prompt specialization on distinct visual regions, while their mean representation covers the most salient parts of the object.}
    \label{fig:heatmaps}
\end{figure*}

\textit{Qualitative Analysis.} Figure~\ref{fig:heatmaps} visualizes the spatial similarity maps between local patch features and learnable local prompt embeddings. These maps demonstrate that OT matching prevents prompt collapse: different prompts specialize on distinct visual regions (e.g., head, tail, eyes), while their average aggregates these specializations to cover the object's most salient parts. This visualization supports our quantitative findings that sparse selection and OT together ground class-specific prompts on complementary, discriminative local features rather than redundantly attending to the same high-scoring regions.

\textit{Summary.}
Our experiments confirm three key findings: (1)~SOT-GLP achieves state-of-the-art average few-shot accuracy by combining global category-level alignment with local part-level specialization; (2)~improvements are most consistent on datasets where textures, parts, or spatial patterns provide discriminative signal; and (3)~the framework offers a controllable accuracy-robustness trade-off via the local projection, enabling deployment-specific tuning.

\section{Ablations on local design choices}

To isolate the contribution of each local branch component, we conduct three targeted ablations while keeping the global branch and training protocol fixed. Table~\ref{tab:nvv-acc-transposed} reports 16-shot accuracy on 11 benchmarks for: (1)~\textbf{SOT-GLP (full)}; (2)~\textbf{w/o V-V attention}, using standard CLIP patch tokens instead of the V-V stream; (3)~\textbf{w/o local projection}, removing the learnable projection before OT; and (4)~\textbf{w/o class-specific locals}, replacing per-class local prompts with a shared pool. 

\textit{Value-value attention.}
Removing V-V attention reduces average accuracy by 0.3\,pp, with the full model winning on 9 out of 11 datasets. V-V attention provides the largest gains on texture-heavy (DTD: +0.2) and scene recognition (SUN397: +0.6, EuroSAT: +1.9) tasks where local patterns dominate. Interestingly, disabling V-V attention slightly improves Aircraft (57.6 → 59.0) and Cars (89.2 → 89.6), suggesting that for these extremely fine-grained tasks, standard CLIP features may already capture relevant global structure, and V-V's emphasis on local interactions can occasionally distract from holistic shape cues. For OOD detection (Table~\ref{tab:ood-fullnames}), both variants achieve similar FPR95 (28.1), but V-V attention improves AUC (93.2 vs.\ 92.9) and substantially benefits texture-based OOD detection (Textures: 31.2 FPR95 vs.\ 34.9). Overall, V-V attention strengthens locality-aware representations that benefit most tasks without degrading global alignment.

\textit{Local projection.}
Removing the learnable projection causes the largest accuracy drop (–0.9\,pp average), with severe degradation on Aircraft (–3.8) and moderate losses on fine-grained datasets (Cars: –0.9, DTD: –0.9). This confirms that task-adaptive projection helps align patch tokens to class-specific prompts, especially when discriminative details are localized. However, this variant achieves \textbf{state-of-the-art OOD detection}: 23.8 FPR95 and 94.2 AUC (Table~\ref{tab:ood-fullnames}), outperforming all baselines. The trade-off is intuitive: learnable projection specializes features to in-distribution classes, improving few-shot accuracy but reducing separation between ID and OOD samples. Freezing the projection preserves CLIP's pretrained geometry, yielding better-calibrated confidence under distribution shift. 

\textit{Class-specific local prompts.}
Replacing class-specific local prompts with a shared pool reduces average accuracy by 0.6\,pp, with concentrated drops on fine-grained tasks: Aircraft (–1.9), Cars (–1.2), UCF101 (–0.8), and DTD (–0.4). Generic object datasets (Caltech101, Flowers102, OxfordPets) show minimal change (±0.1), confirming that class-specific prompts primarily benefit categories distinguished by localized parts or textures. This supports our hypothesis: per-class prompts enable specialization to discriminative attributes (e.g., aircraft tail shapes, car grilles, texture orientations), while shared prompts encourage redundancy and reduce the complementarity of OT assignments. We therefore retain class-specific local prompts in the default configuration.

\textit{Parameter Sensitivity.}
On ImageNet (16-shot), performance peaks at $\lambda = 0.25$ ($75.5\%$) and remains highly stable within the range [0.25, 1.0] with marginal variance ($\leq 0.3\%$), indicating that the framework is robust to hyperparameter selection. However, reducing the local branch's influence excessively ($\lambda \leq 0.125$) causes a performance drop to $74.9\%$, confirming the necessity of strong local supervision. We therefore fix $\lambda = 0.25$ for all experiments. We further analyze the sensitivity to the number of selected patches $K$ on ImageNet (16-shot), accuracy peaks at $K=10$ ($75.5\%$) and remains robust for $K \in [5, 20]$. Performance drops at $K=1$ due to insufficient context and at $K=100$ ($73.8\%$) due to background noise, validating the benefit of sparse alignment.

\textit{Computational Complexity.}
While the dual-branch V-V attention increases inference cost by approximately 24\% (from 44.9 to 58.8 GFLOPs), this is justified by the significant gains in local feature discrimination. OT matching adds negligible overhead as it operates efficiently on the sparse Top-$K$ patches. Notably, on large-scale datasets like ImageNet, most of the training cost is dominated by the text encoder, which must process thousands of class-specific prompts, rather than by the local visual operations.

\textit{Summary.}
Ablations confirm that each component contributes to overall performance: V-V attention provides locality-aware features (+0.3); local projection enables task adaptation (+0.9); and class-specific prompts enable part-level specialization (+0.6). A local projection free variant further improves OOD detection to state-of-the-art levels (23.8 FPR95, 94.2 AUC) at the cost of a 0.9 pp drop in average accuracy, yet notably retains state-of-the-art performance on the majority of individual datasets.

\begin{table}[t]
\centering
\caption{Ablation study on local branch design (16-shot).}
\label{tab:nvv-acc-transposed}

\footnotesize
\sffamily

\setlength{\tabcolsep}{2.2pt}

\renewcommand{\arraystretch}{1.1}
\rowcolors{2}{white}{gray!10}

\newcolumntype{Y}{>{\centering\arraybackslash}X}

\begin{tabularx}{\linewidth}{l *{4}{Y}}
\toprule

\rowcolor{white}
\textbf{Dataset} & 
\makecell{\textbf{SOT-GLP} \\ (Full)} & 
\makecell{w/o \\ V--V Attn} & 
\makecell{w/o \\ Local Proj} & 
\makecell{w/o \\ Class-Spec} \\
\midrule
ImageNet   & \textbf{75.5} & \textbf{75.5} & 75.4 & 75.2 \\
Caltech101 & \textbf{97.4} & 96.9 & 96.8 & 97.0 \\
OxfordPets & \textbf{94.8} & 93.9 & 94.0 & 94.2 \\
Cars       & 89.2 & \textbf{89.6} & 88.3 & 88.0 \\
Flowers102 & \textbf{99.2} & 99.1 & 98.9 & 99.0 \\
Food101    & \textbf{87.8} & 87.6 & 87.1 & 87.7 \\
Aircraft   & 57.6 & \textbf{59.0} & 53.8 & 55.7 \\
SUN397     & \textbf{78.2} & 77.6 & 77.9 & 78.1 \\
DTD        & \textbf{77.1} & 76.9 & 76.2 & 76.7 \\
EuroSAT    & \textbf{91.7} & 89.8 & 90.8 & 91.1 \\
UCF101     & \textbf{87.5} & 87.2 & 87.2 & 86.7 \\
\midrule
\textbf{Average} & \textbf{85.1} & 84.8 & 84.2 & 84.5 \\
\bottomrule
\multicolumn{5}{l}{\textit{Best per row in \textbf{bold}.}} \\
\end{tabularx}
\end{table}

\begin{table}[t]
\centering
\caption{ImageNet Accuracy and OOD Detection Performance.}
\label{tab:ood-fullnames}

\fontsize{7pt}{8.5pt}\selectfont
\sffamily

\setlength{\tabcolsep}{2pt}

\renewcommand{\arraystretch}{1.1} 
\rowcolors{2}{gray!10}{white}

\newcommand{\rot}[1]{\rotatebox{90}{\textbf{#1}}}

\begin{tabular}{l c ccccc}
\toprule
% --- HEADER ROW 1 ---
\rowcolor{white}
 & & 
 \multicolumn{5}{c}{\textbf{OOD Datasets (FPR95$\downarrow$ / AUC$\uparrow$)}} \\[-2pt]
\cmidrule(lr){3-7} 
% --- HEADER ROW 2 ---
\rowcolor{white}
\textbf{Method} & 
\rot{ImageNet (Acc)} & 
\rot{iNaturalist} & 
\rot{SUN} & 
\rot{Places} & 
\rot{Textures} & 
\rot{Average} \\ 
\midrule
% --- DATA ---
CLIP (MCM)  & 66.7 & 30.9 / 94.6 & 37.7 / 92.6 & 44.8 / 89.8 & 57.9 / 86.1 & 42.8 / 90.8 \\
CLIP (GL)   & 66.7 & 15.2 / 96.7 & 30.4 / 93.1 & 38.9 / 89.9 & 57.9 / 83.6 & 35.5 / 90.8 \\
CoOp (MCM)  & 71.7 & 28.0 / 94.4 & 37.0 / 92.3 & 43.0 / 89.7 & 39.3 / 91.2 & 36.8 / 91.9 \\
CoOp (GL)   & 71.7 & 14.6 / 96.6 & 28.5 / 92.7 & 36.5 / 90.0 & 43.1 / 88.0 & 30.7 / 91.8 \\
PLOT        & 72.6 & 15.9 / 96.6 & 33.7 / 92.8 & 38.2 / 91.0 & 39.2 / 90.2 & 31.8 / 92.7 \\
LoCoOp (M)  & 71.5 & 23.1 / 95.5 & 32.7 / 93.4 & 39.9 / 90.6 & 40.2 / 91.3 & 34.0 / 92.7 \\
LoCoOp (GL) & 71.5 & 16.1 / 96.9 & \textbf{23.4} / \textbf{95.1} & 32.9 / \textbf{92.0} & 42.3 / 90.2 & 28.7 / 93.5 \\
PromptSRC   & 73.2 & 20.6 / 95.7 & 30.1 / 93.7 & 38.0 / 91.1 & 46.0 / 89.0 & 33.7 / 92.4 \\
LSN+CoOp    & 72.9 & 23.5 / 95.5 & 29.8 / 93.5 & 36.4 / 90.9 & 38.2 / 89.5 & 32.0 / 92.3 \\
LSN+CoCoOp  & 72.9 & 21.6 / 95.8 & 26.3 / 94.4 & 34.5 / 91.3 & 38.5 / 90.4 & 30.2 / 93.0 \\
GalLoP      & 75.1 & 13.7 / 97.1 & 24.9 / 94.0 & 32.5 / 91.3 & 38.4 / 90.4 & 27.3 / 93.2 \\
\midrule
\textbf{SOT-GLP} & \textbf{75.5} & \textbf{10.4} / \textbf{98.0} & 31.7 / 92.6 & 39.2 / 90.2 & 31.2 / 92.1 & 28.1 / 93.2 \\
\hspace{0.5em} w/o V--V & 75.5 & 11.2 / 97.7 & 29.9 / 92.9 & 37.1 / 90.4 & 34.9 / 90.4 & 28.1 / 92.9 \\
\hspace{0.5em} \textbf{w/o Proj} & 75.4 & 11.2 / 97.8 & 24.8 / 93.9 & \textbf{31.1} / \textbf{92.0} & \textbf{28.0} / \textbf{93.1} & \textbf{23.8} / \textbf{94.2} \\
\bottomrule
\multicolumn{7}{l}{\textit{Best in \textbf{bold}.}} \\
\end{tabular}
\end{table}

\begin{table}[h!]
\centering
\caption{Detailed Few-Shot Performance (\%) across all datasets.}
\label{tab:fewshot_compact_vertical}

\fontsize{6pt}{7pt}\selectfont 
\sffamily

\setlength{\tabcolsep}{1.5pt}

\setlength{\extrarowheight}{0.5pt}
\renewcommand{\arraystretch}{1.05}

\rowcolors{2}{gray!10}{white}

\newcolumntype{Y}{>{\centering\arraybackslash}p{0.6cm}}
\newcolumntype{M}{>{\raggedright\arraybackslash}p{1.4cm}}

\begin{tabularx}{\linewidth}{M *{5}{Y} | *{5}{Y}}
\toprule

\rowcolor{white}
\textbf{Method} & \textbf{1} & \textbf{2} & \textbf{4} & \textbf{8} & \textbf{16} & \textbf{1} & \textbf{2} & \textbf{4} & \textbf{8} & \textbf{16} \\
\midrule

% === BLOCK 1 ===
\rowcolor{white}
 & \multicolumn{5}{c|}{\textbf{ImageNet}} & \multicolumn{5}{c}{\textbf{Caltech101}} \\
CoOp & 66.5 & 67.8 & 69.0 & 70.8 & 71.7 & 79.9 & 89.0 & 92.1 & 93.4 & 95.4 \\
PromptSRC & 68.0 & 70.0 & 71.2 & 72.3 & 73.1 & 93.7 & 94.5 & 95.3 & 95.7 & 96.1 \\
MaPLe & 63.0 & 66.0 & 68.5 & 70.3 & 72.3 & 92.6 & 94.0 & 94.4 & 95.2 & 96.0 \\
GalLoP & 71.0 & 71.8 & 72.5 & 73.8 & 75.1 & 94.7 & 95.1 & 95.5 & 96.0 & 96.7 \\
\textbf{SOT-GLP} & \textbf{71.2} & \textbf{72.1} & \textbf{72.9} & \textbf{74.3} & \textbf{75.5} & \textbf{95.0} & \textbf{95.3} & \textbf{95.7} & \textbf{96.1} & \textbf{97.4} \\
\midrule

% === BLOCK 2 ===
\rowcolor{white}
 & \multicolumn{5}{c|}{\textbf{Oxford Pets}} & \multicolumn{5}{c}{\textbf{Stanford Cars}} \\
CoOp & 44.1 & 58.4 & 71.2 & 78.4 & 85.3 & 67.4 & 70.5 & 74.5 & 79.3 & 83.1 \\
PromptSRC & 92.0 & 92.5 & 93.4 & \textbf{93.5} & 93.7 & 69.4 & 73.4 & 77.1 & 81.0 & 83.8 \\
MaPLe & 89.1 & 90.9 & 91.9 & 92.6 & 92.8 & 66.7 & 71.6 & 75.3 & 79.5 & 83.6 \\
GalLoP & 92.2 & \textbf{93.2} & 93.4 & 93.4 & 94.1 & 71.0 & \textbf{76.1} & 78.3 & \textbf{84.9} & \textbf{89.2} \\
\textbf{SOT-GLP} & \textbf{92.4} & \textbf{93.2} & \textbf{93.7} & 93.4 & \textbf{94.8} & \textbf{71.4} & 74.4 & \textbf{80.0} & 84.3 & \textbf{89.2} \\
\midrule

% === BLOCK 3 ===
\rowcolor{white}
 & \multicolumn{5}{c|}{\textbf{Flowers102}} & \multicolumn{5}{c}{\textbf{Food101}} \\
CoOp & 69.7 & 85.1 & 92.0 & 96.1 & 97.4 & 71.2 & 73.4 & 77.1 & 80.2 & 82.2 \\
PromptSRC & 85.9 & 91.2 & 93.9 & 96.3 & 97.6 & 84.9 & 85.7 & 86.1 & 86.9 & 87.5 \\
MaPLe & 83.3 & 88.9 & 92.7 & 95.8 & 97.0 & 80.5 & 81.5 & 81.8 & 83.6 & 85.3 \\
GalLoP & 84.9 & 92.5 & 95.2 & 98.5 & 98.8 & 83.9 & 84.5 & 85.0 & 86.6 & 87.3 \\
\textbf{SOT-GLP} & \textbf{86.4} & \textbf{93.5} & \textbf{96.6} & \textbf{98.6} & \textbf{99.2} & \textbf{85.9} & \textbf{86.2} & \textbf{86.6} & \textbf{87.3} & \textbf{87.8} \\
\midrule

% === BLOCK 4 ===
\rowcolor{white}
 & \multicolumn{5}{c|}{\textbf{FGVC Aircraft}} & \multicolumn{5}{c}{\textbf{SUN397}} \\
CoOp & 19.6 & 26.4 & 32.3 & 39.4 & 45.4 & 66.8 & 66.5 & 69.9 & 71.5 & 74.7 \\
PromptSRC & 27.7 & 31.7 & 37.5 & 43.3 & 47.1 & 69.7 & 71.6 & 74.0 & 75.7 & 77.2 \\
MaPLe & 26.7 & 30.9 & 34.9 & 42.0 & 48.4 & 64.8 & 67.1 & 70.7 & 73.2 & 75.5 \\
GalLoP & \textbf{32.8} & \textbf{36.2} & \textbf{44.1} & \textbf{51.0} & \textbf{58.3} & 68.3 & 71.6 & 73.8 & 75.7 & 77.2 \\
\textbf{SOT-GLP} & 30.0 & 34.5 & 41.4 & 49.8 & 57.6 & \textbf{71.6} & \textbf{72.9} & \textbf{74.6} & \textbf{76.1} & \textbf{78.2} \\
\midrule

% === BLOCK 5 ===
\rowcolor{white}
 & \multicolumn{5}{c|}{\textbf{DTD}} & \multicolumn{5}{c}{\textbf{EuroSAT}} \\
CoOp & 34.6 & 40.8 & 55.7 & 63.5 & 69.9 & 49.2 & 62.0 & 77.1 & 84.4 & 87.2 \\
PromptSRC & 56.2 & 60.0 & 65.5 & 69.9 & 72.7 & \textbf{73.1} & \textbf{79.4} & \textbf{86.3} & \textbf{88.8} & \textbf{92.4} \\
MaPLe & 52.1 & 55.5 & 61.0 & 66.5 & 71.3 & 71.8 & 78.3 & 84.5 & 87.7 & 92.3 \\
GalLoP & 56.0 & 61.3 & 66.9 & 72.4 & 75.5 & 68.5 & 72.1 & 82.7 & 86.9 & 90.1 \\
\textbf{SOT-GLP} & \textbf{57.0} & \textbf{62.4} & \textbf{67.3} & \textbf{73.5} & \textbf{77.1} & \textbf{70.7} & 74.6 & 83.3 & 88.7 & 91.7 \\
\midrule

% === BLOCK 6 ===
\rowcolor{white}
 & \multicolumn{5}{c|}{\textbf{UCF101}} & \multicolumn{5}{c}{\textbf{Average}} \\
CoOp & 71.2 & 73.4 & 77.1 & 80.2 & 82.2 & 58.2 & 64.8 & 71.6 & 76.1 & 79.5 \\
PromptSRC & 74.8 & 78.5 & 81.6 & 84.3 & 86.5 & 72.3 & 75.3 & 78.3 & 80.7 & 82.9 \\
MaPLe & 71.8 & 74.6 & 78.5 & 81.4 & 85.0 & 69.3 & 72.6 & 75.8 & 78.9 & 81.8 \\
GalLoP & 74.5 & 78.2 & 81.9 & 85.0 & 86.9 & 72.5 & 75.7 & 79.0 & 82.2 & 84.4 \\
\textbf{SOT-GLP} & \textbf{75.4} & \textbf{79.6} & \textbf{82.5} & \textbf{86.2} & \textbf{87.5} & \textbf{73.4} & \textbf{76.3} & \textbf{79.5} & \textbf{82.6} & \textbf{85.1} \\
\bottomrule
\end{tabularx}
\end{table}

\section{Conclusion}
SOT-GLP combines a global prompt branch for category-level semantics with class-specific local prompts matched to salient patches via entropic optimal transport; the dual-branch design further employs V–V attention to produce locality-aware patch features and Top-$K$ selection to suppress background. Empirically, SOT-GLP consistently improves over prior CLIP-based prompt-learning baselines on standard few-shot benchmarks, with the largest gains on tasks where local cues (textures, parts, spatial patterns) complement global semantics. The model is also configurable for OOD detection: removing the learnable local projection yields stronger OOD robustness at a small cost to in-distribution accuracy, allowing practitioners to trade accuracy for robustness as needed. Limitations remain on highly specialized domains where CLIP’s pretrained features lack the required discriminative parts; future work could explore domain-adaptive patch selection, learned Top-$K$ weighting, or alternative local-attention mechanisms to better align prompts with specialized visual cues.

\section*{CRediT authorship contribution statement}
\textbf{Deniz Kizaroğlu:} Conceptualization, Methodology, Software, Writing – original draft, Writing – review \& editing.
\textbf{Ülkü Tuncer Küçüktaş:} Conceptualization, Methodology, Software, Writing – review \& editing.
\textbf{Emre Çakmakyurdu:} Methodology, Visualization, Writing – review \& editing.
\textbf{Alptekin Temizel:} Supervision, Methodology, Writing – review \& editing.

\section*{Data availability}
The datasets used in this study are available in the public domain.

\section*{Declaration of competing interest}
The authors declare that they have no known competing financial interests or personal relationships that could have appeared to influence the work reported in this paper.

\section*{Declaration of generative AI and AI-assisted technologies in the manuscript preparation process}
During the preparation of this work the authors used Gemini in order to improve the language quality and for proofreading. After using this tool/service, the author(s) reviewed and edited the content as needed and take(s) full responsibility for the content of the published article.

\section*{Acknowledgements}
This research did not receive any specific grant from funding agencies in the public, commercial, or not-for-profit sectors.

%% Bibliography
\renewcommand{\bibfont}{\footnotesize}
\bibliographystyle{elsarticle-num}
\bibliography{references}

\end{document}